\pdfoutput=1

\documentclass[11pt]{article}

\usepackage{acl}
\usepackage{times}
\usepackage{latexsym}
\usepackage{multirow}
\usepackage{microtype}
\usepackage{graphicx}
\usepackage{subcaption}
\usepackage{latexsym}
\usepackage{algorithm}
\usepackage{algpseudocode}
\usepackage{booktabs}
\usepackage{courier}
\usepackage{amsmath}
\usepackage{booktabs} 
\usepackage{siunitx} 
\usepackage{placeins}
\usepackage{tikz}
\usepackage[T1]{fontenc}
\usepackage[utf8]{inputenc}
\usepackage{microtype}
\usepackage{inconsolata}
\usepackage{hyperref}

\title{SemEval-2024 Task 2: \\Safe Biomedical Natural Language Inference for Clinical Trials}

\author{\\ \textbf{Maël Jullien\textsuperscript{1}, Marco Valentino\textsuperscript{3}, Andr\'e Freitas\textsuperscript{1}\textsuperscript{,2}\textsuperscript{,3}}\\  $^{1}$Department of Computer Science, University of Manchester, UK\\  
$^{2}$ National Biomarker Centre, CRUK-MI, University of Manchester, UK\\
$^{3}$Idiap Research Institute, Switzerland \\ 
$^{1}${\tt \{firstname.surname\}\tt@manchester.ac.uk}\\
$^{3}${\tt \{firstname.surname\}\tt@idiap.ch}}

\begin{document}
\maketitle
\begin{abstract}

Large Language Models (LLMs) are at the forefront of NLP achievements but fall short in dealing with shortcut learning, factual inconsistency, and vulnerability to adversarial inputs. These shortcomings are especially critical in medical contexts, where they can misrepresent actual model capabilities. Addressing this, we present SemEval-2024 Task 2: Safe Biomedical Natural Language Inference for Clinical Trials. Our contributions include the refined NLI4CT-P dataset (i.e. Natural Language Inference for Clinical Trials - Perturbed), designed to challenge LLMs with interventional and causal reasoning tasks, along with a comprehensive evaluation of methods and results for participant submissions. A total of 106 participants registered for the task contributing to over 1200 individual submissions and 25 system overview papers. This initiative aims to advance the robustness and applicability of NLI models in healthcare, ensuring safer and more dependable AI assistance in clinical decision-making. We anticipate that the dataset, models, and outcomes of this task can support future research in the field of biomedical NLI.  The dataset\footnote{\url{https://github.com/ai-systems/nli4ct}}, competition leaderboard\footnote{\url{https://codalab.lisn.upsaclay.fr/competitions/16190}}, and website\footnote{\url{https://sites.google.com/view/nli4ct/}} are publicly available.

\end{abstract}

\section{Introduction}

\begin{figure}[h!]
\centering
\includegraphics[width=\columnwidth]{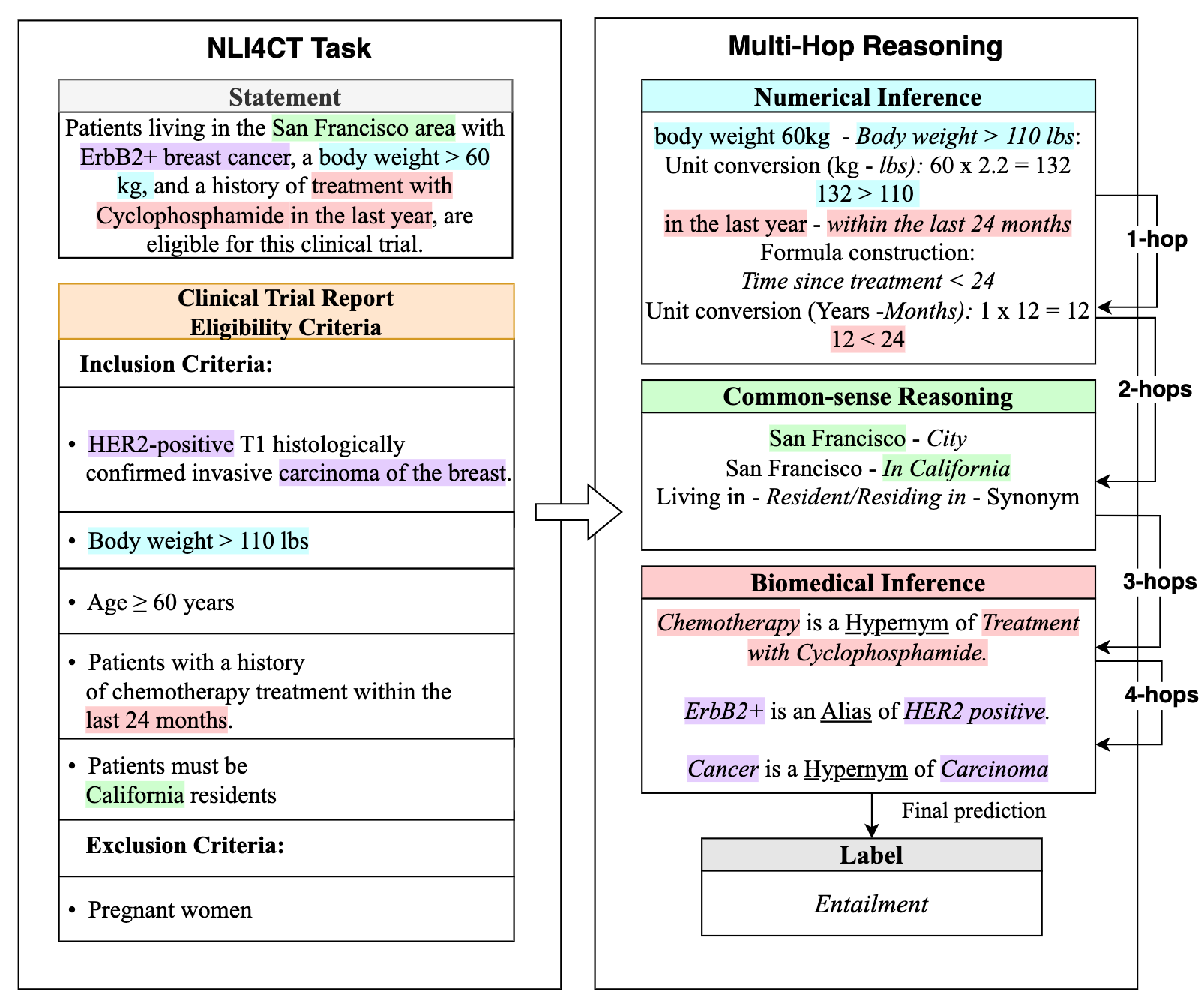}

\caption{The goal of NLI4CT is to predict the relationship of entailment between a \textbf{Statement} and a \textbf{CTR} premise \cite{jullien-etal-2023-nli4ct}. In this task, we introduce a set of perturbations (NLI4CT-P) applied to the statements to test the semantic consistency and faithfulness of NLI models.}
\label{fig:task}
\end{figure}

Large Language Models (LLMs) excel in numerous Natural Language Processing (NLP) tasks, as evidenced by their state-of-the-art achievements \cite{DBLP:journals/corr/abs-2005-14165, chowdhery2022palm}. Despite these advancements, LLMs are prone to several critical vulnerabilities. These include a tendency towards shortcut learning, which may compromise their learning process and accuracy \cite{geirhos2020shortcut, poliak2018hypothesis, tsuchiya2018performance}. Additionally, they exhibit factual inconsistencies \cite{elazar2021measuring} and are sensitive to changes in word distributions \cite{miller2020effect,lee2020biobert}, data transformations \cite{xing2020tasty,stolfo2022causal,meadows2023symbolic,rozanova2023estimating}, and adversarial attacks \cite{li2020bert}. These issues are particularly concerning as they may lead to an overestimation of LLMs' capabilities in practical applications, a risk that is notably significant in fields requiring high reliability, such as healthcare \cite{patel2008investigating, recht2019imagenet}.

Clinical trials play a pivotal role in evaluating the efficacy and safety of novel treatments, thereby significantly contributing to the progress of experimental medicine \cite{avis2006factors}. Clinical Trial Reports (CTRs) document the methodologies and outcomes of these trials, serving as a foundation for healthcare professionals to devise and administer experimental therapies. However, the sheer volume of CTRs, exceeding 400,000 and continually growing \cite{bastian2010seventy}, renders it impractical for a manual comprehensive analysis of all pertinent literature in treatment planning \cite{DeYoung2020EvidenceI2}. In this context, Natural Language Inference (NLI) \cite{bowman2015large} emerges as a viable solution, facilitating the large-scale interpretation and synthesis of medical evidence. This approach effectively bridges the latest research findings with clinical practice, thereby supporting the delivery of personalized care \cite{sutton2020overview}.

Previously, we created the Multi-Evidence Natural Language Inference for Clinical Trial Reports (NLI4CT) dataset, detailed in \citet{jullien-etal-2023-nli4ct}. This dataset, enriched with Clinical Trial Reports (CTRs) and expert-annotated statements for entailment and contradiction, exemplified in Figure \ref{fig:task}, served as the foundation for organizing "SemEval-2023 Task 7: Multi-Evidence Natural Language Inference for Clinical Trial Data".

While the preceding version of NLI4CT spurred the creation of models based on Large Language Models (LLMs) \cite{THiFLY-2023-nli4ct, Saama-2023-nli4ct, Sebis-2023-nli4ct} that demonstrated commendable performance (i.e., F1 score $\approx 85\%$), deploying LLMs in sensitive areas like real-world clinical trials mandates additional scrutiny. This necessitates the invention of new evaluation frameworks that allow thorough behavioural and causal analysis \cite{wang2021measure}.

In pursuit of these goals, we present the latest iteration of our dataset, NLI4CT-P, an extension of the original NLI4CT with data perturbations. Moreover, we provide a comprehensive analysis of the systems that participated in "SemEval-2024 Task 2: Safe Biomedical Natural Language Inference for Clinical Trials" a task conducted using the NLI4CT-P dataset. This initiative aims to improve our understanding of LLMs behaviour and advance evaluation methodologies for clinical Natural Language Inference (NLI). 

\begin{table}[t]
\footnotesize
\begin{minipage}{\columnwidth}
\textbf{Original Statement:} The primary trial intervention protocol lasts a total of 14 days.\\ 
\textbf{Label:} Entailment 
\end{minipage}
\centering
\begin{tabular}{p{0.48\columnwidth}p{0.23\columnwidth}p{0.15\columnwidth}}
\midrule
\textbf{Perturbed Statement} & \textbf{Intervention} & \textbf{Type} \\    \midrule
\raggedright{The primary clinical trial's intervention treatment plan has a duration of 14 days.} & Paraphrase & Preserving \\\midrule
\raggedright{The primary clinical trial intervention protocol spans an entire year} & Contradiction rephrasing & Altering \\\midrule
\raggedright{Lacks energy refers to whether an individual has/had a lack of energy. The primary trial intervention protocol lasts a total of 14 days} & \raggedright{Text appended} & Preserving \\\midrule
\raggedright{The primary trial intervention protocol lasts 2 weeks} & Numerical paraphrase & Preserving \\\midrule
\raggedright{The primary trial intervention protocol lasts a total of 3 hours }& Numerical contradiction & Altering \\ \bottomrule
\end{tabular}
\caption{Example of perturbations applied to the statements with the type of intervention and its semantic effect (i.e., preserving vs. altering).}
\label{tab:inter}
\end{table}

The task is structured around the systematic application of controlled interventions, each designed to investigate specific semantic and numerical inference challenges typical of clinical NLI (see Table \ref{tab:inter}). The interventions enable a comprehensive evaluation of LLMs' reasoning capabilities within a clinical framework, emphasizing robustness, consistency, and faithfulness.

Our efforts aim to significantly contribute to the crafting of more dependable and insightful evaluation standards and metrics for NLI systems, ensuring their reliability and efficacy in healthcare applications.

This second iteration is intended to ground NLI4CT in interventional and causal analyses of NLI models \cite{yu2022interventional}. By enriching the original NLI4CT dataset with a novel contrast set derived from targeted interventions to statements in the NLI4CT test and development sets, we establish a direct causal link between these interventions and the anticipated labels. This enhancement introduces two innovative metrics, Consistency and Faithfulness. These metrics allow us to explore specific research objectives with a causal perspective:

\begin{itemize}
    \item \textbf{Consistency:} To examine whether NLI models maintain uniformity in processing semantically equivalent phenomena crucial for inference within clinical NLI contexts.
    \item \textbf{Faithfulness:} To assess the capacity of NLI models to capture and interpret the underlying semantic features required for reasoning over clinical trials, and to change their predictions according to relevant changes of such features.
\end{itemize}

This paper introduces SemEval-2024 Task 2 -- Safe Biomedical Natural Language Inference for Clinical Trials -- (NLI4CT-P) presenting a detailed analysis of the performance of the participating systems. We report the following conclusions:

\paragraph{Challenges in Clinical NLI:} Despite improvements achieved via the application of Large Language Models (LLMs), Clinical NLI remains a significant challenge. With the highest F1 score achieved in this task being 0.8 \cite{liu-thoma-2024-fzi-wim, guimaraes-etal-2024-lisbon} (FZI-WIM, Lisbon Computational Linguists), leveraging  Mixtral-8x7B-Instruct models. This emphasises the necessity for the development of more robust and reliable systems capable of dealing with the challenges of real-world clinical application.

\paragraph{Importance of Faithfulness and Consistency Evaluation:} The incorporation of Faithfulness and Consistency metrics into our evaluation framework underscores the unpredictability of current systems and the limitations inherent in relying solely on F1 score for comprehensive analysis.

\paragraph{Superiority of Generative Models:} Generative models have been shown to outperform discriminative models in terms of F1 score (+0.025), Faithfulness (+0.15), and Consistency (+0.037).

\paragraph{Value of Additional Data:} Leveraging additional training data in the form of instruction tuning or medical NLI datasets has been shown to produce significant performance gains. When augmented with extra data, systems exhibit notable enhancements, recording improvements of +0.056 in F1 score, +0.132 in Faithfulness, and +0.062 in Consistency relative to their counterparts.

\paragraph{Impact of Prompting Strategies:} The study highlights that the choice of prompting strategy plays a crucial role in influencing model performance. Specifically, zero-shot prompting has been shown to provide notable enhancements, with an average increase of +0.025 in F1 score, and marginal gains of +0.001 in both Faithfulness and Consistency, compared to the outcomes achieved with few-shot prompting techniques. 

\paragraph{Efficacy of Mid-Sized Architectures:} Mid-sized architectures, possessing 7B to 70B parameters, offer a cost-effective alternative capable of matching or surpassing larger models in key performance metrics like F1, Faithfulness, and Consistency. Compared to models exceeding 70B parameters, these mid-sized models report a slight improvement of +0.01 in F1 score, albeit with minor reductions of -0.03 in Faithfulness and -0.01 in Consistency. Against models below 7B parameters, however, they show notable enhancements, achieving +0.10 in F1 score, +0.40 in Faithfulness, and +0.19 in Consistency.

\section{Task Description}

SemEval-2024 Task 2 is a textual entailment task, each instance in NLI4CT contains a CTR premise and a related statement. These premises range from 5 to 500 tokens in length and provide details about a trial's results, eligibility criteria, interventions, or adverse events. Corresponding statements are concise sentences, containing 10 to 35 tokens, that make some claim about the premise information (refer to Table \ref{tab:inter} for examples). The task is to classify the inference relation between a CTR premise, and a statement as either entailment or contradiction, exemplified in Figure \ref{fig:task} The dataset features two distinct types of instances: single instances, where a statement discusses a single CTR, and comparison instances, which involve statements that compare and contrast two CTRs.

\section{Dataset}

The premises used in the NLI4CT dataset \cite{jullien-etal-2023-nli4ct} are derived from 1,000 publicly accessible, English-language breast cancer Clinical Trial Reports (CTRs) published on \href{https://clinicaltrials.gov/ct2/home}{ClinicalTrials.gov} a resource managed by the U.S. National Library of Medicine. This dataset complies with the Health Insurance Portability and Accountability Act (HIPAA) Privacy Rule. The original NLI4CT collection includes 2,400 expert-annotated statements, premises and associated labels. These are distributed across training, testing, and development sets in a 70/20/10 ratio.

We have advanced the methodology of the previous NLI4CT dataset by incorporating interventions to create a contrast set, enabling a systematic behavioural and causal analysis of models evaluated in the competition. This enhanced version is referred to as NLI4CT-P (Perturbed). The construction of the contrast set involves four semi-automated, controlled interventions applied to the statements from the NLI4CT test and development set. It's important to note that the specifics of these interventions were kept undisclosed until the completion of the competition's testing phase on January 31st 2024. 

\subsection{Interventions}
We delineate and implement the four interventions in the following manner:
\paragraph{Paraphrasing and Contradiction Rephrasing} 
Clinical texts frequently contain acronyms and aliases, which can hinder the performance of clinical NLI models \cite{grossman2021deep,jimeno2011exploiting,pesaranghader2019deepbiowsd,jin2019deep}. Moreover, these models can fall prey to shortcut learning, where they make inferences based on syntactic patterns rather than semantic understanding \cite{geirhos2020shortcut}. To evaluate this phenomenon, original statements were rephrased using different vocabulary and syntax. Paraphrasing was employed to retain the original meaning and label (row 1 Table \ref{tab:inter}), while contradiction rephrasing created new statements that directly contradict the original statement, and are therefore always labelled as contradictions (row 2 Table \ref{tab:inter}).

\paragraph{Numerical Paraphrasing and Contradiction}
Large Language Models (LLMs) have shown limitations in consistent numerical and quantitative reasoning  \cite{DBLP:journals/corr/abs-2103-07191,DBLP:journals/corr/abs-1901-03735,DBLP:journals/corr/abs-1903-11907}, an essential aspect for tasks like NLI4CT that demand such inferences. To evaluate the models' capabilities in this area, operands and numerical units within the hypotheses were altered (rows 4 and 5 Table \ref{tab:inter}). This modification either preserved or inverted the initial entailment label. 

\paragraph{Appending Text} LLMs are often challenged by complex reasoning when dealing with extended premise-hypothesis pairs \cite{liu2021natural}. We test this in a clinical setting by appending biomedical definitions from the \href{https://ncithesaurus.nci.nih.gov/ncitbrowser/}{NCI Thesaurus} to the original statements (row 3 Table \ref{tab:inter}). The added definitions, ranging from 15 to 20 tokens in length, almost double the average statement token length. Despite the definitions not being independently verifiable against the premises, these definitions are regarded as 'ground truth', they are universally true and remain neutral in relation to the premises. Since they neither assert nor verify any premise-specific information, within the scope of our task, appending such neutral text is categorized as a 'preserving' intervention.

\begin{table*}[h]
\footnotesize
\centering
\begin{tabular}{p{0.05\columnwidth}p{0.14\columnwidth}p{0.2\columnwidth}p{0.2\columnwidth}p{0.22\columnwidth}p{0.25\columnwidth}p{0.25\columnwidth}p{0.1\columnwidth}}
\toprule
\textbf{Set}& \textbf{Original} & \textbf{Appended definition}  & \textbf{Paraphrase} & \textbf{Contradiction rephrasing} & \textbf{Numerical paraphrase} & \textbf{Numerical Contradiction} & \textbf{Total} \\    \midrule
\textbf{Dev} & 200 & 600 & 600 & 600 & 64 & 78 & 1942 \\\midrule
\textbf{Test} & 500 & 1500 & 1500 & 1500 & 224 & 276 & 5000 \\
\bottomrule
\end{tabular}
\caption{Distribution of statement counts across the sets of NLI4CT-P}
\label{tab:my_label}
\end{table*}

These interventions, other than the text appending, were performed by prompting ChatGPT 3.5 and Whisper APIs \cite{openai} with human-in-the-loop correction to address any errors \cite{gilardi2023chatgpt}. Each statement in the test and development sets underwent each type of intervention process three times. This did not extend to the training set, as the aim was to prevent models from learning the patterns of intervention. Although attempts were made to apply numerical paraphrasing and contradiction interventions, they were not always feasible. This was due to the absence of numerical data or units in the original statements, and when the quality of the perturbed statements was deemed substandard, they were excluded during the manual review phase. consequently, this resulted in a markedly reduced count of numerically perturbed statements within the dataset. The prompts used to perform the interventions are available in the appendix.

\section{Evaluation}
SemEval-2024 Task 2 is devised as a binary classification challenge, with the Macro F1-score being utilized to gauge the foundational performance of the participating systems. This evaluation is conducted on the original NLI4CT test set, serving as a control metric, rather than on the NLI4CT-P test set, which contains exclusively perturbed statements. Although the Macro F1 score is instrumental in measuring overall model performance by highlighting precision and recall across various classes, it inherently lacks the capability to fully capture the sophisticated understanding and reasoning skills essential for effective Natural Language Inference (NLI). Specifically, the F1 score does not assess a model's capacity to adjust to subtle semantic shifts or evaluate the resilience of its predictions when faced with interventions that either modify or maintain the semantic integrity of statements. This gap highlights the necessity for more advanced metrics capable of offering deeper insights into a model's interpretative and reasoning proficiency. In response to this need and inspired by recent advancements in causal analysis within the NLP domain \cite{stolfo2022causal}, we introduce two novel evaluation metrics aimed at examining the causal effects of interventions on model performance.

\paragraph{Faithfulness} gauges the degree to which a system's predictions are both accurate and grounded in the correct rationale. Intuitively, this is estimated by measuring the ability of a model to correctly adjust its predictions when exposed to interventions that modify the meaning (semantic altering) of the statement. Specifically, for a set of $N$ statements $x_i$ in the contrast set ($C$), alongside their corresponding original statements $y_i$ and the model predictions denoted as $f()$, faithfulness is quantified using the formula presented in Equation 1.

\begin{equation}
\footnotesize
    \begin{aligned}
   Faithfulness = \frac{1}{N}\sum_{1}^{N}\left| f(y_i)-f(x_i) \right|\hspace{1cm}\\ x_i\in C:\text{Label}(x_i) \neq \text{Label}(y_i), \text{ and } f(y_i) = \text{Label}(y_i) 
    \end{aligned}
\end{equation}

\paragraph{Consistency} assesses a system's capability to generate identical outcomes for semantically equivalent inputs. This measure evaluates whether a system can uniformly predict the same label for both the original and contrast statements under interventions that do not alter the semantic content (semantic preserving) of the statements. The key aspect here is the uniformity in representing semantic concepts across different statements, irrespective of the correctness of the final prediction. For $N$ statements $x_i$ in the contrast set ($C$), alongside their original counterparts $y_i$, and model predictions $f()$, consistency is determined as follows:

\begin{equation}
\footnotesize
    \begin{aligned}
   Consistency = \frac{1}{N}\sum_{1}^{N} 1 - \left| f(y_i)-f(x_i) \right| \\ x_i\in C:\text{Label}(x_i) = \text{Label}(y_i)
    \end{aligned}
\end{equation}

The Macro F1 score provides a foundational benchmark for basic model performance, serving as a control metric, the core objective of Task 2 is towards enhancing model quality and dependability through systematic causal analysis. The pursuit here is not only for high performance in a traditional sense but for models that demonstrate a more reliable and robust application of natural language, reflecting a more nuanced approach to evaluating system capabilities, and allowing for developing safer, ethical, and trustworthy clinical systems.

\begin{table*}[h!]
\centering
\tiny 
\begin{tabular}{@{}p{2.5cm}p{0.3cm}p{0.3cm}p{0.3cm}p{0.5cm}p{2cm}p{1.8cm}p{1cm}p{3.5cm}}
\toprule
\textbf{Work } & \textbf{F1} & \textbf{F} & \textbf{C} & \textbf{Average Score} & \textbf{Architecture} & \textbf{Inference Strategies} & \textbf{Fine-Tuning} & \textbf{Dataset Augmentation} \\ \midrule
FZI-WIM \cite{liu-thoma-2024-fzi-wim} & \textbf{0.8} & 0.9 & 0.73 & 0.81 & Mixtral-8x7B-Instruct & CoT & Yes & GPT-4, bart-large-mnli Instruction Dataset \\\midrule
Lisbon Computational Linguists \cite{guimaraes-etal-2024-lisbon} & \textbf{0.8} & 0.83 & 0.72 & 0.78 & Mistral-7B-Instruct-v0.2 & Zero-shot & Yes &  Mistral-7B-Instruct-v0.2 dataset expansion \\\midrule
NYCU-NLP \cite{lee-etal-2024-nycunlp} & 0.78 &  0.92 &  \textbf{0.81} & \textbf{0.84} & SOLAR (10.7B) & Zero-shot & Yes & OpenChat v3.5, Intervention Reduction \\\midrule
Edinburgh Clinical NLP \cite{hong-etal-2024-edinburgh}	&0.78	&\textbf{0.95}&	0.78	&\textbf{0.84}	&GPT-4&	Zero-shot	& No &- \\\midrule
YNU-HPCC \cite{zhang-etal-2024-ynu-hpcc} & 0.77 & 0.67 & 0.73 & 0.72 & DeBERTa-v3-large & Discriminative & Yes & MultiNLI, FeverNLI, ANLI, LingNLI, WANLI, Back Translation \\\midrule
BD-NLP \cite{nath-samin-2024-bdnlp} & 0.77 & 0.79 & 0.76 & 0.77 & DeBERTa-lg & Discriminative & Yes  & - \\\midrule
CaresAI	\cite{abdel-salam-etal-2024-caresai}&0.77	&0.76	&0.75	&0.76	&Ensemble of DeBERTas	&Discriminative	&Yes  & - \\\midrule
TüDuo \cite{smilga-alabiad-2024-tuduo} & 0.76 & 0.84 & 0.75 & 0.78 & Flan-T5 XL & Few-shot & Yes &  GPT-3.5-Turbo Instruction Dataset \\\midrule
RGAT \cite{chakraborty-2024-rgat} & 0.76 & 0.86 & 0.74 & 0.79 & GPT-4 & Zero-shot & No  & - \\\midrule
DFKI-NLP \cite{verma-raithel-2024-dfki-nlp} & 0.75 & 0.81 & 0.68 & 0.75 & Mistral 7B & Zero-shot & Yes &  Meta-Inventory dataset expansion, MedNLI \\\midrule
D-NLP \cite{altinok2024evaluating} & 0.75 & 0.83 & 0.74 & 0.77 & Gemini Pro & Zero-shot & No & - \\ \midrule
LMU-BioNLP \cite{sun-etal-2024-lmubionlp} & 0.75 & 0.86 & 0.69 & 0.77 & Mistral-7b & Zero-shot & Yes & GPT-3.5, GPT4 dataset expansion, and instruction tuning dataset \\\midrule
DKE-Research \cite{wang-etal-2024-dkeresearch} & 0.74 & 0.8 & 0.75 & 0.76 & DeBERTa-l & Discriminative & Yes& GPT-3.5, TF-IDF dataset expansion \\\midrule
Puer \cite{dao-etal-2024-puer} & 0.72 & 0.59 & 0.64 & 0.65 & Biolinkbert-large & Discriminative & Yes  & - \\\midrule
UniBuc \cite{micluta-campeanu-etal-2024-unibuc} & 0.71 & 0.83 & 0.72 & 0.75 & SOLAR 10B & few-shot & No  & - \\\midrule
iML \cite{akkasi-etal-2024-iml} & 0.7 & 0.28 & 0.52 & 0.50 & SciFive & Zero-shot & Yes & - \\\midrule
CRCL \cite{brutti-mairesse-2024-simple} & 0.7 & 0.87 & 0.7 & 0.76 & Mixtral-8x7B & CoT, OPRO optimization & No  & - \\\midrule
IITK \cite{mandal-modi-2024-iitk} & 0.69 & 0.9 & 0.71 & 0.77 & Gemini Pro & Zero-shot, ToT and CoT & No  & - \\\midrule
0x.Yuan \cite{lu-kao-2024-0xyuan} & 0.68 & 0.51 & 0.56 & 0.58 & Mixtral-8x7B & multi-agent debating framework & No  & - \\\midrule
Saama Technologies \cite{kim-etal-2024-saamatech} & 0.66 & 0.59 & 0.58 & 0.61 & Gemini Pro, mistral-7B-instruct-v0.2 & CoT, Few-Shot & Yes  & - \\\midrule
TLDR \cite{das-etal-2024-tldr} & 0.66 & 0.5 & 0.58 & 0.58 & SciFive-base, DeBERTa-v3-base & Zero-shot & No & -  \\\midrule
Concordia University \cite{marks-etal-2024-faithful}&	0.66&	0.03&	0.39&	0.36&	BART&	Discriminative&	Yes	&-\\\midrule
T5-Medical \cite{siino-2024-t5medical} & 0.63 & 0.3 & 0.5 & 0.48 & T5-large-medical & Zero-Shot & No & -  \\\midrule
USMBA-NLP \cite{fahfouh-etal-2024-bert} & 0.62 & 0.44 & 0.54 & 0.53 & BERT base & Discriminative & Yes & -  \\\midrule
SEME \cite{aguiar-etal-2024-seme} & 0.57 & 0.64 & 0.56 & 0.59 & NLI-RoBERTa ensemble & Discriminative & Yes & -  \\
\hline
\end{tabular}
\caption{SemEval-2024 Task 2 Results, sorted by F1 (on the unperturbed subset of the test set), with Faithfulness (F), and Consistency (C)}
\label{tab:semeval2024}
\end{table*}

\section{Results and Discussion}

106 participants registered to the SemEval-2024 Task 2 competition contributing over 1200 individual submissions and 25 system overview papers, presented in Table \ref{tab:semeval2024}. Please note that our analysis focuses exclusively on systems that are detailed in system overview papers and for which official leaderboard results have been provided. Generally, participants tend to submit the highest-scoring results to the leaderboard, regardless of whether the system achieving these results represents the primary contribution of their paper. This approach ensures that our report reflects the peak performance levels achieved, albeit potentially overlooking the main systems of interest described in the papers.

\subsection{Architectures}
\begin{figure}[t]
  \centering
  \includegraphics[width=\linewidth]{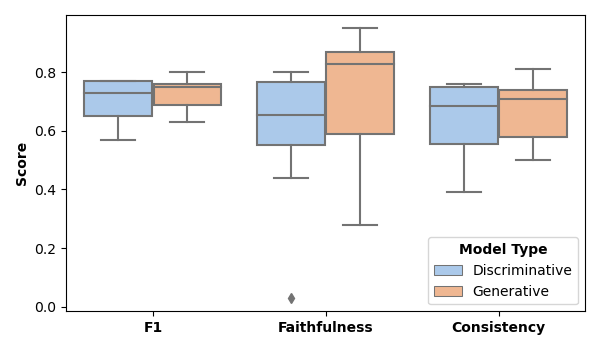}
  \caption{Comparative Analysis of F1, Consistency, and Faithfulness Across Model Types}
\label{fig:model types}
\end{figure}

\begin{figure}[t]
  \centering
  \includegraphics[width=\linewidth]{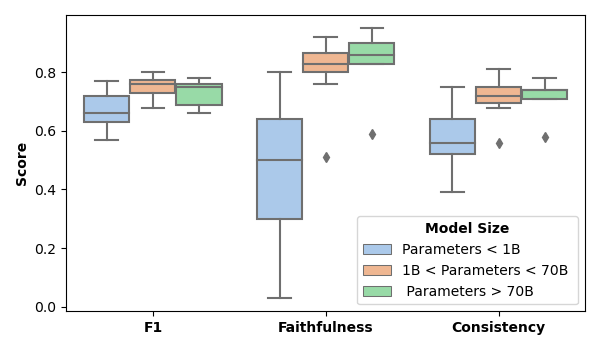}
  \caption{Comparative Analysis of F1, Consistency, and Faithfulness Across Model Parameter Numbers}
\label{fig:param}
\end{figure}

\begin{table}[t]
\centering
\caption{Participant architectures by popularity, with average F1, Faithfulness (F) and Consistency (C)}
\label{table:model_usage}
\begin{tabular}{llllc}
\toprule
\textbf{Model} & \textbf{F1} & \textbf{F} & \textbf{C} &\textbf{Count} \\
\midrule
DeBERTa &0.76	&0.76&	0.75& 5 \\
Mistral 7B& 0.75&	0.84	&0.69& 4 \\
Mixtral 8x7B& 0.73	&0.76	&0.66& 3 \\
T5& 0.66&	0.36&	0.53& 3 \\
Gemini Pro &0.70&	0.77&	0.68& 3 \\
GPT-4 &\textbf{0.77}	&\textbf{0.91}&	0.76& 2 \\
SOLAR 10B &0.75&	0.88	&\textbf{0.77}& 2 \\
BERT base& 0.62	&0.44&	0.54& 1 \\
Biolinkbert& 0.72	&0.59&	0.64& 1 \\
BART &0.66	&0.03	&0.39& 1 \\
RoBERTa& 0.57	&0.64	&0.56& 1 \\
Flan-T5 XL&0.76&	0.84&	0.75 & 1
\\\bottomrule
\label{tab: models}
\end{tabular}
\end{table}

In the SemEval-2024 Task 2 submissions, a diverse range of 12 different architectures was employed, as detailed in Table \ref{tab: models}. The predominant choice among participants was Mistral-based architectures, accounting for 7 out of 25 submissions, closely followed by DeBERTa with 5 out of 25. The majority of submissions utilised generative models, with 17 out of the total, compared to 8 leveraging discriminative models. The F1 score suggests that GPT-4's performance is on par with considerably smaller models such as DeBERTa. However, a deeper evaluation using our novel metrics, especially Faithfulness, reveals a significant disparity, indicating that smaller models might be overfitting. This observation underscores the importance of employing these complementary metrics for a more comprehensive comparison of model capabilities. Despite the prevailing notion that larger models inherently perform better, this trend appears to be less pronounced than observed in this task's previous iteration \cite{jullien-etal-2023-semeval}, as illustrated in Figure \ref{fig:param}. Notably, there seems to be a point of diminishing returns for model sizes between 7B and 70B, within the generative model category, shown in Figure \ref{fig:param}. On average, models with sizes ranging from 7B to 70B parameters achieve +0.01 in F1 score but show decreases of -0.03 in Faithfulness and -0.01 in Consistency relative to models with more than 70B parameters. When compared to models with fewer than 7B parameters, these mid-sized models exhibit substantial improvements of +0.10 in F1 score, +0.40 in Faithfulness, and +0.19 in Consistency.

Additionally, on average, generative models outperform discriminative ones across the board—with improvements observed in F1 scores (+0.025), Faithfulness (+0.15), and Consistency (+0.037), as depicted in Figure \ref{fig:model types}. Intriguingly, when comparing specific architectures, there is minimal correlation between model types and Faithfulness, Consistency, and F1, even though the top two performing systems in terms of F1 score are based on the Mixtral-8x7B-Instruct model (see Table \ref{tab:semeval2024}).

\subsection{Base F1 Performance}

As previously mentioned the focus of this task extends beyond base performance. Nevertheless, it's noteworthy that the highest F1 score achieved in this iteration was 0.8 \cite{liu-thoma-2024-fzi-wim, guimaraes-etal-2024-lisbon} (FZI-WIM, Lisbon Computational Linguists) by two systems (Table \ref{tab:semeval2024}). A figure that notably falls short of the previous iteration's top score of 0.856 \cite{THiFLY-2023-nli4ct, jullien-etal-2023-semeval}. This observed decline underscores a significant gap between the current capabilities of NLI systems and the performance required for practical application within clinical environments.

\subsection{Faithfulness and Consistency}

The overall average Faithfulness recorded at 0.719 significantly outperforms the average Consistency, which stands at 0.67. This disparity grows more pronounced within the subset of models within the top 10 F1 scores, where Average Faithfulness escalates to 0.835 and Average Consistency to 0.751.

Furthermore, a robust overall Spearman's correlation was identified between Consistency and F1 scores (0.8) and between Faithfulness and F1 scores (0.62). Intriguingly, this correlation inverts within the top 10 systems, where Spearman's Correlation between Consistency and F1 drops to -0.12, and between Faithfulness and F1 rises slightly to 0.319. Notably, the models with the highest Faithfulness (0.95) \cite{hong-etal-2024-edinburgh}(Edinburgh Clinical NLP) and Consistency (0.81) \cite{lee-etal-2024-nycunlp}(NYCU-NLP) scores achieve an average score of 0.84, surpassing systems ranked above them (with average scores of 0.81 and 0.78) yet both reporting a lower F1 score by -0.02. also \citet{mandal-modi-2024-iitk}(IITK) achieves a very high faithfulness of 0.9, while only managing an F1 of 0.69. These patterns underscore the limitation of F1 scores as sole indicators of model performance at the apex levels, accentuating the importance of considering Faithfulness and Consistency metrics in conjunction with F1.

The inversion of correlations among the top 10 models suggests a nuanced landscape of performance evaluation. While Consistency contributes broadly to high F1 scores, the top 10 models distinctly leverage Faithfulness, indicating that, at peak performance levels, perhaps accurate predictions rooted in correct premises are paramount over consistent responses to similar cases.

This phenomenon might also signify a ceiling effect for Consistency, suggesting that beyond a certain point, efforts to improve consistency do not translate into proportional performance gains. Such a scenario could inadvertently overshadow other critical model attributes like adaptability and nuanced comprehension, aspects more closely associated with Faithfulness. Alternatively, this situation could imply that models specifically optimized for F1 scores might inadvertently neglect Consistency, and to some degree, Faithfulness, as evidenced by the observed decline in their correlation with peak F1 scores.

Our analysis further elucidates the relationship between Consistency and Faithfulness in submitted systems, revealing an Overall Spearman Correlation of 0.708. This correlation slightly diminishes among the top 10 F1 scoring models to 0.39. While this represents a weaker correlation within the subset of the top 10 models, it importantly suggests the absence of a strict trade-off between Consistency and Faithfulness. Such a finding challenges the notion that improvements in one metric necessarily come at the expense of the other.

Among the participants, 4 out of 25 achieved a Faithfulness score of 0.9 or higher \cite{mandal-modi-2024-iitk,liu-thoma-2024-fzi-wim, lee-etal-2024-nycunlp, hong-etal-2024-edinburgh}(IITK, FZI-WIM, NYCU-NLP, Edinburgh Clinical NLP). Remarkably, only 1 out of 25 participants attained a Consistency score of 0.8 or higher \cite{lee-etal-2024-nycunlp}(NYCU-NLP). These results suggest a continued need for refining these models to achieve higher degrees of Faithfulness and Consistency if they are to be applied in real-world clinical environments.





\subsection{Prompting Strategies}
\begin{figure}[t]
  \centering
  \includegraphics[width=\linewidth]{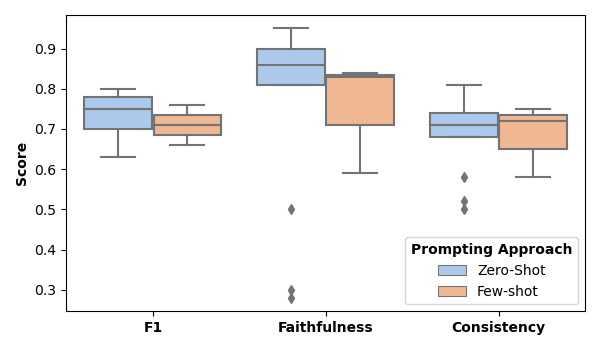}
  \caption{Comparative Analysis of F1, Consistency, and Faithfulness Across Prompting strategies}
\label{fig:prompt}
\end{figure}

A variety of prompting strategies were used in the submitted systems. It is essential to acknowledge that variations in prompts can lead to significant differences in outcomes, even when employing the same architecture. For instance, within the Gemini Pro systems, a comparison between submissions by \citet{altinok2024evaluating}(D-NLP) and \citet{kim-etal-2024-saamatech}(Saama Technologies) from Saama Technologies reveals substantial disparities in performance metrics: F1 scores, Faithfulness, and Consistency differ by 0.09, 0.24, and 0.16, respectively. Similar patterns of variation were observed among submissions utilizing Mistral-based and T5-based approaches, underscoring the impact of prompting nuances.

Among the generative model submissions, 13 out of 16 employed a zero-shot approach, while the remaining three opted for few-shot prompting. Zero-shot prompting involves generating responses without any example-based guidance, relying solely on the model's pre-existing knowledge and the task description. Few-shot prompting, on the other hand, provides the model with one or more examples to guide its responses, traditionally anticipated to yield superior results.

Contrary to initial expectations, zero-shot prompting has shown a significant advantage, especially in achieving higher F1 scores and improving Faithfulness. Notably, four out of the top five models with the highest F1 scores utilized zero-shot techniques, as depicted in Figure \ref{fig:prompt}. On average, zero-shot prompting yielded improvements of +0.025 in F1 score, +0.001 in Faithfulness, and +0.001 in Consistency, when compared to few-shot prompting methods. 

Direct prompting is a straightforward method of querying a Language Model (LM). It involves posing a question to the model in a direct manner, without providing additional context or requesting intermediate steps. For example \textit{"Given the CTR: }\{Premise\} \textit{does the statement: }\{Statement\} \textit{follow?"}

On the other hand, Chain of Thought (CoT) prompting represents a more elaborate technique designed to prompt the model to "show its work" by articulating the intermediate steps or reasoning that leads to its conclusion \cite{DBLP:journals/corr/abs-2201-11903}. This approach enables the model to break down the problem into smaller, more manageable parts, thereby facilitating more accurate or explainable predictions. For instance, the prompt could be structured as follows: \textit{"Given the CTR: }\{Premise\} \textit{and the statement:} \{Statement\}\textit{, provide a step-by-step reasoning process to determine if the statement logically follows from the report."} Such a modification in the prompting strategy has been shown to produce significant differences in the model's outputs \cite{DBLP:journals/corr/abs-2201-11903}.

While direct prompting has been the predominant strategy among generative approaches, several teams have experimented with more nuanced strategies. Specifically, FZI-WIM \cite{liu-thoma-2024-fzi-wim}, IITK \cite{mandal-modi-2024-iitk}, and Saama Technologies \cite{kim-etal-2024-saamatech} have employed Chain of Thought prompting. Furthermore, IITK \cite{mandal-modi-2024-iitk} has also explored Tree of Thought (ToT) prompting. ToT prompting is an advanced technique aimed at improving the performance and interpretability of LMs, particularly in complex problem-solving tasks \cite{yao2023tree}. It goes beyond the CoT approach by not merely listing reasoning steps linearly but by organizing these steps into a tree structure that represents different branches of reasoning or possible solutions. IITK \cite{mandal-modi-2024-iitk} applies this technique with the prompt \textit{Imagine three different clinical experts are answering the question given below. All experts will write down first step of their thinking, then share it with the group. Then all experts will go on to the next step of their thinking. If any expert realises they're wrong at any point then they leave. They will continue till a definite conclusion is reached.}. However, the ability to draw definitive conclusions about the relative efficacy of these prompting strategies is constrained given the considerable performance variability associated with each approach and the application of these strategies across diverse models, complicating efforts to ascertain the sources of performance gains or losses.

Two particularly intriguing prompting strategies emerged from the submissions. \cite{brutti-mairesse-2024-simple}(CRCL) utilized an OPRO (Optimal Prompting for Response Optimization) technique \cite{yang2023large}, which leverages the model's ability to generate effective prompts from a small set of exemplars and prior instructions. This technique essentially tasks the model with creating its own instructions to tackle given problems. Additionally, \cite{lu-kao-2024-0xyuan} introduced a multi-agent debating framework, incorporating several custom agents with diverse expertise, including Biostatistics and Medical Linguistics, to enrich the model's output.

In summary, the submissions reveal a broad spectrum of prompting strategies, from zero-shot to more complex approaches like Tree of Thought and multi-agent frameworks. These strategies significantly influence model performance, underscoring the importance of prompt design in the development and evaluation of NLI systems. As the field progresses, further research is warranted to elucidate the optimal prompting strategies for enhancing model accuracy, reliability, and interpretability across various applications, in a controlled manner.

\subsection{Fine-tuning strategies}
\begin{figure}[t]
  \centering
  \includegraphics[width=\linewidth]{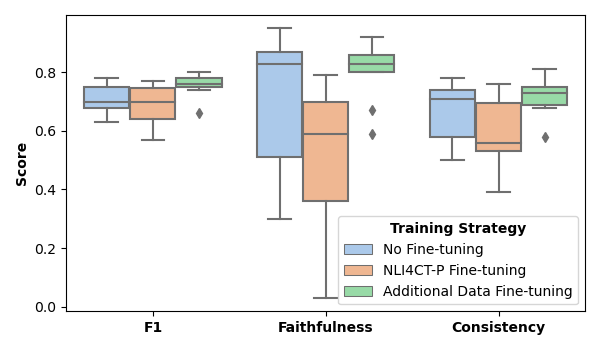}
  \caption{Comparative Analysis of F1, Consistency, and Faithfulness Across Training Strategies}
\label{fig:finetune}
\end{figure}

Within the context of SemEval-2024 Task 2, a diverse array of fine-tuning strategies was employed across the 25 participating systems, revealing significant insights into their impact on model performance. Notably, 9 out of 25 systems, all of which were generative, did not undergo any form of fine-tuning. In contrast, 8 out of 25 systems were fine-tuned specifically on the NLI4CT-P training set, while the remaining 6 systems benefited from fine-tuning on additional datasets.

Interestingly, systems fine-tuned on the NLI4CT-P training set exhibited the lowest average performance across all three evaluated metrics, as detailed in Figure \ref{fig:finetune}. Conversely, systems that underwent fine-tuning on external datasets demonstrated superior performance on all metrics, indicating a significant advantage of incorporating diverse training data.

The range of additional datasets leveraged for fine-tuning included various medical NLI datasets, such as MultiNLI, FeverNLI, ANLI, LingNLI, and WANLI, utilized by \citet{zhang-etal-2024-ynu-hpcc}(YNU-HPCC), and MedNLI by \citet{verma-raithel-2024-dfki-nlp}(DFKI-NLP). Moreover, some teams, including \citet{sun-etal-2024-lmubionlp}(LMU-BioNLP), \citet{wang-etal-2024-dkeresearch}(DKE-Research), \citet{guimaraes-etal-2024-lisbon}(Lisbon Computational Lin-
guists), \citet{smilga-alabiad-2024-tuduo}(TüDuo), and \citet{zhang-etal-2024-ynu-hpcc}(YNU-HPCC), innovatively generated their data by applying interventions similar to those used in our task, thereby enriching their training material. Systems enhanced with additional data demonstrate significant improvements, achieving gains of +0.056 in F1 score, +0.132 in Faithfulness, and +0.062 in Consistency. These results suggest a substantial benefit from such tuning, particularly in terms of Faithfulness. This indicates that incorporating perturbed data into the training process not only enhances the model's inference ability but also significantly improves its reliability and adherence to the truthfulness of the clinical data it processes.

Instruction tuning emerged as a prevalent strategy, with datasets specifically crafted for this purpose by teams such as \citet{liu-thoma-2024-fzi-wim}(FZI-WIM), \citet{guimaraes-etal-2024-lisbon}(Lisbon Computational Lin-
guists), \citet{smilga-alabiad-2024-tuduo}(TüDuo), LUM-BIO, \citet{wang-etal-2024-dkeresearch}(DKE-Research), and \cite{lee-etal-2024-nycunlp}(NYCU-NLP). Notably, 3 out of the top 5 systems, as per F1 scores, employed instruction tuning, underscoring its effectiveness in enhancing model performance, although notably producing minimal gains in consistency.


\section{Related Work}
The landscape of expert-annotated resources for clinical NLP is rich, with notable examples such as the TREC 2021 Clinical Track \cite{soboroff2021overview}, which focuses on information retrieval from CTR data, highlighting eligibility criteria. Evidence Inference 2.0 \cite{DeYoung2020EvidenceI2} introduces a QA task alongside span selection based on CTR results, while the MEDNLI dataset \cite{romanov2018lessons} offers an entailment task using patient medical history notes. These datasets primarily aim to evaluate biomedical language understanding and reasoning. Despite neural architectures leading in biomedical NLI performance \cite{gu2021domain, DeYoung2020EvidenceI2}, challenges remain in quantitative reasoning and numerical operations within NLI \cite{DBLP:journals/corr/abs-1901-03735,DBLP:journals/corr/abs-1903-11907}. Prior works experiment with biomedical pre-training strategies \cite{lee2020biobert, shin2020biomegatron, gu2021domain}, and while ExaCT \cite{kiritchenko2010exact} automates information extraction from clinical trials, the integration of biomedical and numerical NLI effectively remains unaddressed. None of the aforementioned resources provide avenues for meaningful causal analysis, a gap NLI4CT-P aims to fill, through the application of targeted interventions and the introduction of novel evaluation metrics.

\section{Conclusion}

This study introduces the NLI4CT-P dataset and provides a comprehensive analysis of submissions to SemEval-2024 Task 2, underscoring the persistent challenges in Clinical Natural Language Inference (NLI) despite significant advancements in Large Language Models (LLMs). The incorporation of Faithfulness and Consistency metrics further highlights these challenges, shedding light on areas requiring additional focus, if these systems are to meet the requirements for real-world clinical implementation. Our key findings reveal that generative models markedly outperform discriminative models, particularly in terms of Faithfulness and Consistency. The utility of additional data is underscored, especially due to the limited size of the NLI4CT-P training set. Furthermore, our analysis reveals the substantial impact of prompting strategies on model performance, noting an intriguing preference for zero-shot approaches over few-shot methods. Additionally, mid-sized architectures, ranging between 7B and 70B parameters, demonstrate the potential to match or even exceed the performance of larger models (>70B) in F1 scores, Faithfulness, and Consistency, while being more resource and cost-effective. Conversely, models with fewer than 7B parameters face difficulties in achieving comparable results. We plan to perform a further analysis of the submitted systems' performance at an intervention level, identifying specific areas of weakness, such as numerical reasoning or handling longer premises, to refine and enhance Clinical NLI systems further.

\section{Limitations}

Despite not disclosing detailed specifics of the interventions, nor providing intervened training data, several participants generated their own interventions for data augmentation. As a result, some models were specifically trained on this intervened data. However, this approach raises concerns regarding their ability to generalize effectively to entirely new, unseen perturbations or adversarial datasets. The tailored training to specific interventions may limit the models' broader applicability and robustness on unseen perturbed or adversarial data.

\section{Acknowledgments}
This work was partially funded by the Swiss National Science Foundation (SNSF) project NeuMath (\href{https://data.snf.ch/grants/grant/204617}{200021\_204617}), by the EPSRC grant EP/T026995/1 entitled “EnnCore: End-to-End Conceptual Guarding of Neural Architectures” under Security for all in an AI enabled society, by the CRUK National Biomarker Centre, and supported by the Manchester Experimental Cancer Medicine Centre.

\bibliography{custom}

\appendix
\newpage
\section{Intervention Prompts}

\subsection{Contradictory Rephrasing prompt}
\begin{quote}
\itshape
Your task is to provide 3 contradictory statements, given an original statement.

(Instructions)
 Ensure that the contradictory statements are factually opposed to the original statement. 
 Do not mention the original statement in the contradictory statements. 
 Use formal and straightforward language when writing the new statements, and avoid unusual or overly descriptive language. 
 Make sure to retain the names 'Primary Clinical Trial' and 'Secondary Clinical Trial' in the contradictory statements, these names must be present in every statement.
 Provide 3 different options in a consistent JSON format with keys 'Statement\_1', 'Statement\_2', and 'Statement\_3' followed by their respective paraphrased statements. 

(Examples)
1. [original statement]:"the secondary trial requires patients to be over a certain age, but the primary trial does not specify an age range for participation."
[ideal output]: "the secondary trial does not give an age limit for patients to participate, but patients must be between the age of 12-34 to be eligible for the primary trial"

2. [original statement]:"a patient that has received an organ transplant within the last month, and is still bedridden would be excluded from the primary trial but may be eligible for the secondary trial"
[ideal output]: "a patient that has received an liver transplant in the last week, with an ECOG score of 4 would be eligible for the primary trial but excluded from the secondary trial"

3.[original statement]: "Women with Newly diagnosed stage IV breast cancer, confirmed as ER+ Considering a mastectomy are eligible for the primary trial"
[ideal output]: "Women recently diagnosed with stage 4 ER-positive breast cancer and contemplating a mastectomy are excluded from the Primary Clinical Trial"

Input: 
\end{quote}
\subsection{Paraphrasing prompt}
\begin{quote}
\itshape
Your task is to provide 3 paraphrased statements, given an original statement.

(Instructions)
 Use formal and straighforward language when writing the new statements, and avoiding unusual or overly descriptive language. 
 Make sure to retain the name 'Primary Clinical Trial' in the statements, this name must be present in every statement. 
 Provide 3 different options in a consistent JSON format with keys 'Statement\_1', 'Statement\_2', and 'Statement\_3' followed by their respective paraphrased statements. 

(Examples)
1. [original statement]:"the primary trial does not specify an age range for participation."
[ideal output]: "patients aged between 30-60 years old can be eligible for the primary trial"

2. [original statement]:"a patient that has received an organ transplant within the last month, and is still bedridden would be excluded from the primary trial"
[ideal output]: "a patient that has received an liver transplant in the last week, with an ECOG score of 4 would be excluded from the primary trial"

3.[original statement]: "Women with Newly diagnosed stage IV breast cancer, confirmed as ER+  Considering a mastectomy are eligible for the primary trial"
[ideal output]: "Women recently diagnosed with stage 4 ER-positive breast cancer and contemplating a mastectomy are suitable for the Primary Clinical Trial"

Input: 

\end{quote}
\subsection{Numerical Paraphrasing prompt}
\begin{quote}
\itshape
Your task is to modify the numerical values and units in an original statement while maintaining its original meaning, to generate 3 new statements.

(Instructions)
Do not paraphrase the statements, You can only change numerical values or units, if you change the units you must also convert the measurement values.
Provide 3 different options in a consistent JSON format with keys 'Statement\_1', 'Statement\_2', and 'Statement\_3' followed by their respective paraphrased statements. 

(Examples)
1. [original statement]: "Over 6 weeks of TAK-228 Plus Tamoxifen treatment patients in the primary trial experienced a 5\% reduction in the Percentage of cells with Ki67 expression"
[ideal output]: "Over 42 days of TAK-228 Plus Tamoxifen treatment patients in the primary trial experienced a 5\% reduction in the Percentage of cells with Ki67 expression"

2.[original statement]: "in the primary trial there were 10 times the number of Hepatotoxicity cases as there were cases of hypertension and Pancreatectomy"
[ideal output]: "in the primary trial there were 1000\% the number of Hepatotoxicity cases as there were cases of hypertension and Pancreatectomy"

3.[original statement]: "2/73 the primary trial participants, and 0/1674 the secondary trial participants suffered an Acute myocardial infarction "
[ideal output]: "2.74\% the primary trial participants, and 0\% the secondary trial participants suffered an Acute myocardial infarction "

Input: 

\end{quote}
\subsection{Numerical Contradictory Rephrasing prompt}
\begin{quote}
\itshape
Your task is to modify the numerical values and units in an original statement to contradict the original statement, to generate 3 new statements.

(Instructions)
Do not paraphrase the statements, You can only change numerical values or units, if you change the units you must also convert the measurement values.
Provide 3 different options in a consistent JSON format with keys 'Statement\_1', 'Statement\_2', and 'Statement\_3' followed by their respective paraphrased statements. 

(Examples)
1. [original statement]: "Over 6 weeks of TAK-228 Plus Tamoxifen treatment patients in the primary trial experienced a 5\% reduction in the Percentage of cells with Ki67 expression"
[ideal output]: "Over 50 days of TAK-228 Plus Tamoxifen treatment patients in the primary trial experienced a 105\% reduction in the Percentage of cells with Ki67 expression"

2.[original statement]: "in the primary trial there were 10 times the number of Hepatotoxicity cases as there were cases of hypertension and Pancreatectomy"
[ideal output]: "in the primary trial there were 30\% the number of Hepatotoxicity cases as there were cases of hypertension and Pancreatectomy"

3.[original statement]: "2/73 the primary trial participants, and 0/1674 the secondary trial participants suffered an Acute myocardial infarction "
[ideal output]: "9.74\% the primary trial participants, and 8\% the secondary trial participants suffered an Acute myocardial infarction "

Input: 
\end{quote}

\end{document}